\documentclass[letterpaper]{article} 
\usepackage{aaai25}  
\usepackage{times}  
\usepackage{helvet}  
\usepackage{courier}  
\usepackage[hyphens]{url}  
\usepackage{graphicx} 
\usepackage{natbib}  
\usepackage{caption} 
\usepackage{algorithm}
\usepackage{algorithmic}

\usepackage{newfloat}
\usepackage{listings}
\DeclareCaptionStyle{ruled}{labelfont=normalfont,labelsep=colon,strut=off} 
\floatstyle{ruled}
\newfloat{listing}{tb}{lst}{}
\floatname{listing}{Listing}
\usepackage{tabularx}

\title{Effective Defect Detection Using Instance Segmentation for NDI}
\author{
    Ashiqur Rahman\textsuperscript{\rm 1}, 
    Venkata Devesh Reddy Seethi\textsuperscript{\rm 1},
    Austin Yunker\textsuperscript{\rm 2},
    Zachary Kral\textsuperscript{\rm 3},
    Rajkumar Kettimuthu\textsuperscript{\rm 2}, and
    Hamed Alhoori\textsuperscript{\rm 1}
}
\affiliations{
    \textsuperscript{\rm 1}Department of Computer Science, 
    Northern Illinois University, DeKalb, IL 60115 USA\\

    \textsuperscript{\rm 2}Argonne National Laboratory, Lemont, IL, USA\\

    \textsuperscript{\rm 3}Spirit AeroSystems, Wichita, KS, USA\\
    
    ashiqur.r@niu.edu
}

\begin{document}

\maketitle

\begin{abstract}
Ultrasonic testing is a common Non-Destructive Inspection (NDI) method used in aerospace manufacturing. However, the complexity and size of the ultrasonic scans make it challenging to identify defects through visual inspection or machine learning models. Using computer vision techniques to identify defects from ultrasonic scans is an evolving research area. In this study, we used instance segmentation to identify the presence of defects in the ultrasonic scan images of composite panels that are representative of real components manufactured in aerospace. We used two models based on Mask-RCNN (Detectron $2$) and YOLO $11$ respectively. Additionally, we implemented a simple statistical pre-processing technique that reduces the burden of requiring custom-tailored pre-processing techniques. Our study demonstrates the feasibility and effectiveness of using instance segmentation in the NDI pipeline by significantly reducing data pre-processing time, inspection time, and overall costs.
\end{abstract}

%

\section{Introduction}\label{introduction}
Non-Destructive Inspection (NDI) is inspecting an object without damaging it. This process is essential in manufacturing, construction, defect evaluation, art restoration, infrastructure safety, and many more \citep{Honarvar2020-tw, Ould_Naffa2002-ab, Dwivedi2018-ts, Memmolo2015-er, Brosnan2004-yd}. With the advent of artificial intelligence (AI) and machine learning (ML), automation of the NDI process has become at the forefront of research \citep{Gardner2020-bn, Lin2023-fd}. In this study, we worked with ultrasonic scans of thin-walled composite materials, used in the aerospace manufacturing industry to identify defects in the manufactured product. A trained inspector inspects the scan using proprietary software to identify any defects in the scans. In our work, we want to augment the process by suggesting areas of interest for the inspector to look for defects. With the help of ML, we can reduce the inspection time and inspector fatigue and increase the safety of the manufactured parts.

The process of using ML to identify defects from ultrasonic scans is of great interest to the NDI research community. Several studies suggested methods of identifying defects using ML \citep{Baumgartl2020-fe, Yang2022-nl}. However, the lack of high-quality training data and high training time are constraints to the progress of this field. While researchers agree that ML is the direction to choose for the future of NDI \citep{Jodhani2023-mn, Sun2023-yp}, there are still many improvements to make before it becomes the primary choice \citep{Hassani2023-hi}. In the manufacturing industry, the high cost of deployment and maintenance is also a prohibitive factor. So, for ML-based approaches to have a place in the NDI pipeline, the training and maintenance of the models need to be cheaper, and the method should have very high accuracy to ensure the safety of the critical parts. In this work, we propose image segmentation models as a feasible approach for ML-based NDI systems. While image segmentation is used by many researchers as a tool to pre-process the data, or as a part of the defect detection models, using image segmentation as a viable defect detection approach needs further research.

Object detection using image segmentation models has recently become a rising research area. The object detection models classify images at the pixel level to identify the presence of an object \citep{Hafiz2020-id}. In this study, we used trained two different models - Detectron $2$ based on Mask-RCNN \citep{wu2019detectron2} and ``You Only Look Once" (YOLO) 11 \citep{Redmon2016-ap, Khanam2024-hp} to detect defects in the ultrasonic scans and compared the results to propose this as a viable approach for NDI.

\section{Related Works}\label{relatedworks}
\subsubsection{NDI. } 
NDI is an essential part of the manufacturing pipeline that ensures the quality of manufactured products through inspection without destroying or breaking the products \citep{Gholizadeh2016-ww, Honarvar2020-tw}. This inspection process ensures the reliability of the product right from the start. Additionally, in safety-critical manufacturing domains, such as aerospace, NDI is necessary for the scheduled inspections that are conducted when the product is in use, to ensure there is no major damage occurring from wear and tear \citep{Khedmatgozar_Dolati2021-as}.

\subsubsection{NDI Techniques. }
NDI is an undeniable part of manufacturing, the food industry, structural health monitoring, inspecting artifacts, and many more. In the airline industry, the inspection of composite material is an essential part of NDI process. Studies such as \citet{Diamanti2005-mr, Gholizadeh2016-ww, Dwivedi2018-ts, Honarvar2020-tw} showed different effective techniques of inspection, including but not limited to, Visual Testing, Ultrasonic Testing, Thermography, Radiographic Testing, Electromagnetic Testing, Acoustic Emission, Shearography Testing, and so on. With the advancement of technology over time, composite structures are preferred over metals for their lightweight and durability in the aerospace industry. Metal parts used eddy current and magnetic particle induction techniques for NDI \citep{Lange1994-nf}. However, these techniques do not work on composite structures like carbon fiber-reinforced polymer (CFPR) materials. Advances in NDI techniques addressed this in studies from \citet{Gupta2022-id, Gholizadeh2016-ww}. Ultrasonic testing is the most common method used in the industry \citep{Honarvar2020-tw}. Despite its inability to detect very small defects and dependence on the experience of the inspectors \citep{Gupta2022-id}, ultrasonic testing, due to its ability to detect sub-surface defects and relatively low cost, has been the most popular method for the NDI process \citep{Dwivedi2018-ts, Wang2020-up}. The ultrasonic-based NDI approach requires processing complex signals to interpret for the presence of flaws in a structure. Due to this complexity, the use of ML has become inevitable.

\subsubsection{ML in NDI. } 
Many research articles evaluated \citep{Mishra2020-qv} and discussed the importance, usefulness, and challenges \citep{Gardner2020-bn, Taheri2022-ix, Lin2023-fd} of ML in NDI.

\citet{Mishra2020-qv} evaluated artificial neural networks (ANN), adaptive neuro-fuzzy inference systems (ANFIS), and support vector regression (SVR) models and found SVR most promising. While discussing the undeniable performance improvement of NDI with the help of ML. \citet{Gardner2020-bn} also underscored the challenges of data availability, data quality, and complexity in training the ML models. Other methods like gradient boosting decision tree (GBDT) \citep{Yang2022-nl}, principal component analysis (PCA), and support vector machine (SVM) \citep{Ma2020-cu} are proposed for NDI techniques. However, appropriate hyper-parameter tuning and quality training data are critical for the better performance of ML \citep{Sun2023-yp}. 
 
\subsubsection{Image Segmentation. }
Image segmentation is an incremental research area that has evolved over decades and is rooted in classification research. The instance segmentation techniques, which have become highly effective in recent times, combine the detection of object bounding boxes and categorically detecting every image pixel \citep{Hafiz2020-id}. Image segmentation on ultrasonic images is a crucial technique in the computer-aided diagnosis (CAD) field, where researchers use semantic segmentation techniques to identify cancerous areas in the ultrasonic images \citep{Su2011-re, Irfan2021-ra}. A similar technique can be used in ultrasonic scan images of other materials to identify defects. Several deep learning models have also been proposed by the research community like UPSNet, Panoptic-DeepLab, Detectron, and so on \cite{Kirillov2019-uc, Xiong2019-qk, Cheng2020-sa, Wu2019-iw}. The available models use medical or real-world object datasets for training \citep{Elharrouss2021-wa}. In aerospace, several works have employed semantic segmentation for identifying aircraft components such as engine, wing, and fuselage using DeepLab $v3$ \citep{thomas2024advanced} and YOLO $v5$ \citep{xiang2024aircraft}. Furthermore, computer vision was also used for identifying surface defects on aircraft from drone images using YOLO-FDD \citep{li2024yolo}. However, identifying defects from $3D$ ultrasonic images is challenging due to the images requiring additional pre-processing to transform into $2D$ images. \citet{prakash2023learning} have used a software to first visualize the fuselage scan, apply complex feature processing techniques such as histogram of gradients, and train a KNN classifier to identify defects.

In this study, we aim to use simple pre-processing techniques on raw $3D$ ultrasonic scans and train instance segmentation models to identify defects. Furthermore, we do a comparative analysis of two popular segmentation frameworks such as Detectron $2$ and YOLO $11$, and evaluate the computational needs as well as their performance. Developing instance segmentation models that can achieve desirable performance with minimal pre-processing, enhances the scalability of the method as well as reduce the time and computational cost. 

\section{Data}\label{data}

We used a proprietary dataset containing scans of specimen panels, which are thin-walled composite structures representative of aerospace structures used in the industry. The panels consist of multiple plies of unidirectional carbon fiber polymer. In some of the specimens, an additional fiberglass ply was added on top of the layup. The specimen has a range of thicknesses to account for changes in actual structure. Teflon strips were placed throughout the layup at controlled locations to represent defects in the material. Per industry standards, the Teflon strips at varying depths and locations. The Teflon material has a sound impedance that can represent foreign inclusions or delaminations in the composite structure.

\begin{figure}[h]
    \centering
    \includegraphics[width=\linewidth]{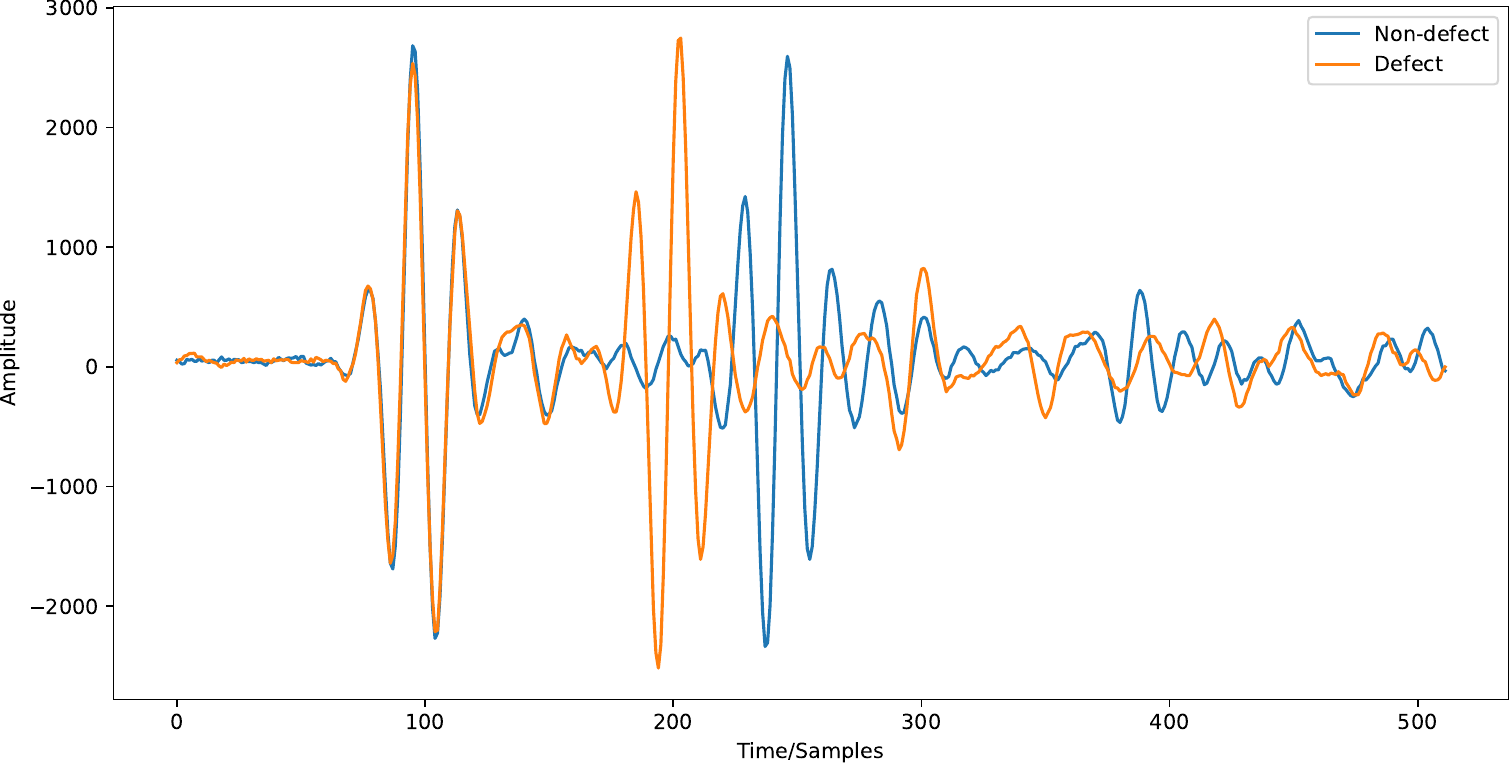}
    \caption{Sample of signals showing the difference between defect and non-defect areas} 
    \label{fig:figure1}
\end{figure}

A single transducer system with $2.5MHz$ and $5MHz$ frequencies is used to scan the plates from both sides. The transducer was positioned perpendicular to the structure and sent ultrasonic waves through the material, and a receiver at the opposite of the object received the signals. From each point of scans, $512$, $1024$, and $2048$ samples were collected for further analysis. Fluctuations in the amplitude of the received signals indicate the presence of defects. Differences in the signal attenuation can be observed from the visualization in Figure \ref{fig:figure1}.

Each scan comprising the signal data has height and width dimensions of $258\times368$ and channels $512$, $1024$, or $2048$, depending on sampling frequency as discussed above. Since we use 2D segmentation models, we first convert these $3D$ scans to $2D$ data by taking the variance over the sampling frequency dimension. Then, we converted the $2D$ data into \textit{NumPy} arrays and exported them as \textit{PNG} images. For the purpose of training the models, we converted the defect annotations into COCO formatted \citep{Lin2014-jw} JSON files. The annotated JSON contained the bounding boxes for each defect which will be used for the training. To reduce the resource usage during training, we converted the images into grayscale. Using this process, we created $72$ files for our study. We randomly selected $56$ images for training, $8$ for validation, and $8$ for testing our models. Figure \ref{fig:figure2} (A) shows an example of the exported images used for training. Orange labels in Figure \ref{fig:figure2} (B) indicate the defect regions on the training image.

\section{Method}\label{method}
We used two different models to identify the defects in the exported images. We used the instance segmentation approach, which is a more refined version of semantic segmentation, where each pixel of an image is classified, and a cluster of pixels is classified as part of a class \cite{Hafiz2020-id}. We used two different pre-trained algorithms, Detectron $2$ based on Mask-RCNN \citep{wu2019detectron2} and YOLO $11$ \citep{Jocher2023-xo}, and fine-tuned them to train on our datasets.

\subsection{Mask R-CNN (Detectron $2$)}
We used the instance segmentation feature of the Detectron $2$ library, which utilizes the Mask R-CNN model. In this approach, the model first detects an object and creates a bounding box. Subsequently, it classifies the pixels inside that bounding box. To optimize model performance, we trained the model using various configurations and hyperparameter settings, such as number of epochs, batch size, images per batch, etc.

\subsubsection{Training Configuration. }
To train the Detectron $2$ model, we prepared the COCO dataset discussed above. The training was performed on a machine with $12GB$ GPU and $64GB$ RAM. To find the optimal model configuration, we conducted a series of training experiments using various fine-tuned hyperparameters, as summarized in Table \ref{tab:det2_configuration}. The performance of each model was measured using the mean average precision (mAP) metric at the intersection over union (IOU) \citep{Rahman2016-uu} threshold of $0.5$ and $0.75$. To maintain consistency and ensure efficient processing, all images were resized to a standardized $512\times512$ pixel format. No additional image augmentation techniques were employed during the training process.

\begin{table}[h]
    \small
    \centering
    \begin{tabularx}{\linewidth}{
    >{\raggedright\arraybackslash}X
    >{\raggedleft\arraybackslash}X
    >{\raggedleft\arraybackslash}X
    >{\raggedleft\arraybackslash}X
    >{\raggedleft\arraybackslash}X
    >{\raggedleft\arraybackslash}X
    }
        \hline
        Name & Batch Size & Epochs & $mAP^{50}$ & $mAP^{75}$ & Time (hours) \\
        \hline
         d2-10k & $8$ & $10,000$ & $43.56\%$ & $5.30\%$ & $9.52$ \\
         \hline
         d2-50k & $8$ & $50,000$ & $80.60\%$ & $51.16\%$ & $12.22$ \\
         \hline
         d2-90k & $8$ & $90,000$ & $82.65\%$ & $46.96\%$ & $22.50$ \\
         \hline
    \end{tabularx}
    \caption{Detectron $2$ training configurations}
    \label{tab:det2_configuration}
\end{table}

From this experiment, we found that \textit{d2-50k} offered the optimal result offering the best performance while minimizing the training time.

\begin{figure}[h]
    \centering
    \includegraphics[width=\linewidth]{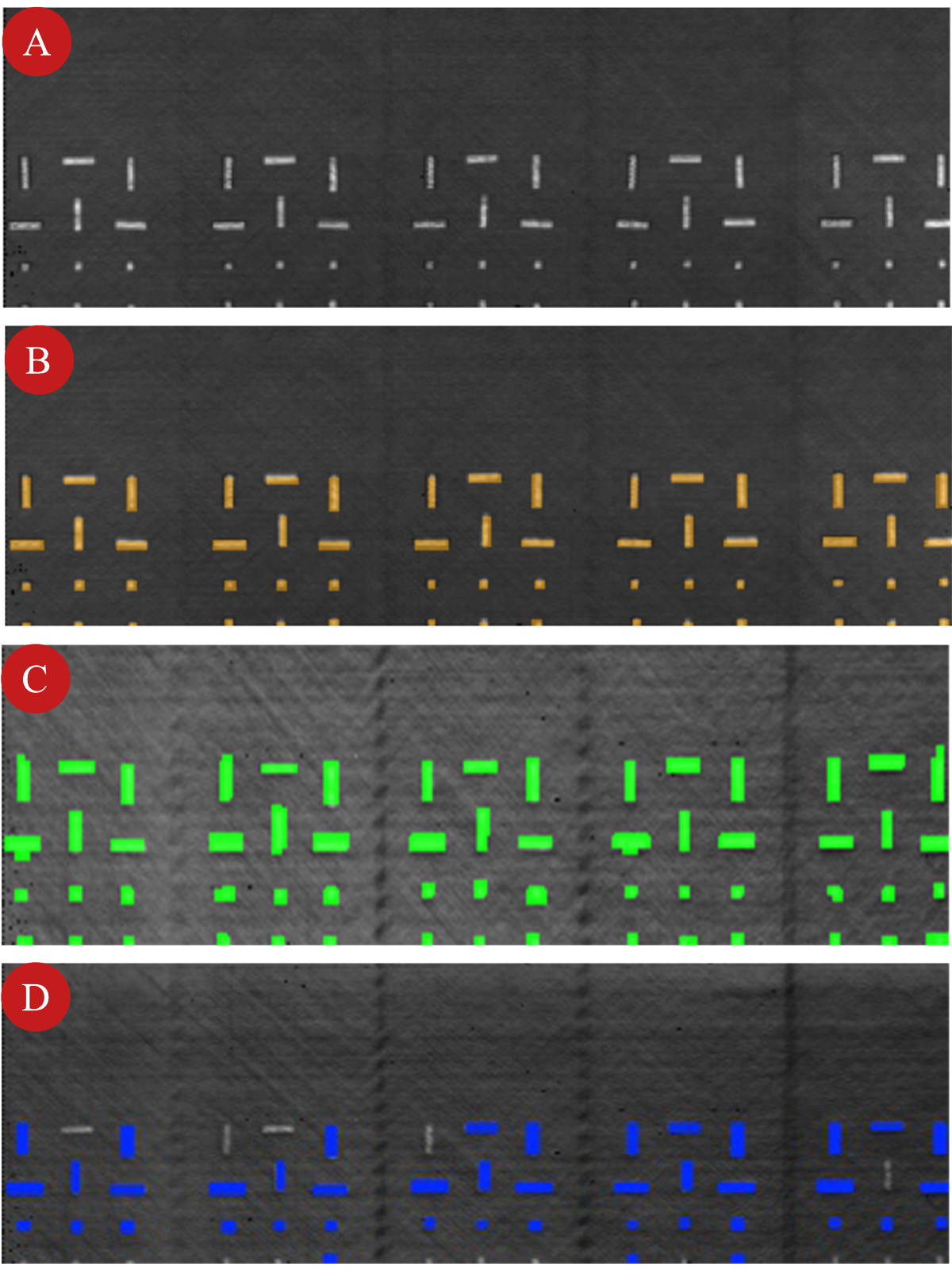}
    \caption{Exported images of the ultrasonic scans. A) The exported image used for the training. B) Orange labels showing the defect location on the image. C) Defects identified by the Detectron $2$ model displayed with green labels. D) Defects identified by the YOLO $11$ model displayed with blue labels.}
    \label{fig:figure2}
\end{figure}

\subsubsection{Results. }
In our experiment, we achieved an average precision exceeding 80\% at the 50\% IOU threshold. However, given the critical safety implications of NDI applications, prioritizing sensitivity over precision is crucial. In this context, false positives are preferable to false negatives, as identifying every potential defect is essential. Since the primary goal of the model is to assist inspectors in locating regions of interest, detecting a fragment of a defect is sufficient to warrant a manual inspection of the affected area.

Figure \ref{fig:figure2} (C) visually represents the ability of the model to accurately identify all potential defect areas within a scan, highlighted by green overlays.

\subsection{YOLO v11}
We used the object detection models of YOLO $11$, which integrates a backbone network for extracting features, followed by a segmentation head that generates both bounding boxes and detailed pixel-level masks for individual objects.

\subsubsection{Training Configuration. }
To train the YOLO $11$ model, we converted the COCO formatted data into appropriate labels and image files accepted by the YOLO \citep{Ultralytics2023-lc} framework. We fine-tuned different hyperparameters of the pre-trained model to find the optimal configuration. We performed the training on the same hardware configuration as the Detectron $2$ experiment. The performance of the training configurations is summarized in Table \ref{tab:yolo11_configuration}. To ensure efficient processing, all images were resized to standardized $640\times640$ pixels during training. We experimented with two pre-trained YOLO $11$ models -- \textit{yolo11n}, a smallest model with $2.6 million$ parameters, and \textit{yolo11x}, a larger model with $56.9 million$ parameters \citep{Ultralytics2024-em}. The training processes were terminated when the model performance converged. The first experiment with $1,000$ epochs converged at $837$ epochs whereas the other experiments converged at $1296$ epochs.

\begin{table}[h]
    \small
    \centering
    \begin{tabularx}{\linewidth}{
    >{\raggedright\arraybackslash}X
    >{\raggedleft\arraybackslash}X
    >{\raggedleft\arraybackslash}X
    >{\raggedleft\arraybackslash}X
    >{\raggedleft\arraybackslash}X
    >{\raggedleft\arraybackslash}X
    }
        \hline
        Name & Batch Size & Max Epochs & $mAP^{50}$ & $mAP^{75}$ & Time (min) \\
        \hline
         \mbox{yolo11n-1k} & $8$ & $1,000$ & $77.04\%$ & $48.41\%$ & $16.75$ \\
         \hline
         \mbox{yolo11n-10k} & $8$ & $10,000$ & $75.22\%$ & $48.66\%$ & $29.55$ \\
         \hline
         \mbox{yolo11x-10k} & $8$ & $10,000$ & $75.22\%$ & $48.66\%$ & $28.68$ \\
         \hline
    \end{tabularx}
    \caption{YOLO $11$ training configurations}
    \label{tab:yolo11_configuration}
\end{table}

\subsubsection{Results. }
From the YOLO $11$ training, we observed that the smaller pre-trained model, \textit{yolo11n}, was sufficient to achieve optimal performance with minimal training time. Similar to the Detectron $2$ experiment, the performance metric, while not perfect, was sufficient to identify almost all the defect areas, as illustrated by the blue overlays in Figure \ref{fig:figure2} (D).

\section{Discussion}\label{discussion}
Our experiments demonstrated that the proposed method significantly reduces data pre-processing requirements. By converting scan data to PNG format, we successfully trained image segmentation models without requiring any complex pre-processing and data normalization. This streamlined pipeline enhanced the practicality of the models for industrial deployment by reducing computational overhead and increasing the model's adaptability across different scanning configurations.

The minimized pre-processing time enables real-time defect detection, allowing inspectors to identify defects as soon as the scanned data is available. This boosts efficiency and strengthens the NDI process. This minimized pre-processing step increased the model's viability across varied industrial environments.

Our experimental results indicate that the YOLO $11$ model exhibited significantly faster training times while achieving performance comparable to the Detectron $2$ model. This accelerated training process is particularly advantageous for rapid prototyping and iterative model development.

Compared to traditional ML models, which often require extensive signal preprocessing and cleanup, image segmentation models demonstrated remarkable performance directly from the raw scan data. This reduced reliance on complex pre-processing techniques simplifies the overall training and deployment pipeline for different manufactured materials.

Additionally, the high detection accuracy and efficiency of these models suggest their potential for practical application in real-world scenarios. The ability to accurately segment objects within images can be leveraged in a wide range of NDI processes.

Since the models rely on PNG images rather than underlying signals, they can easily adapt to defect detection on any new scan from a similarly shaped object. This flexibility minimizes the need for extensive retraining when introducing new manufactured parts. The retraining for new parts can also be simplified by creating images of the new components and incorporating them into the existing training data. This incremental approach allows for efficient adaptation to evolving product designs without requiring a complete retraining process.

\section{Conclusion}\label{conclusion}

Our experimental results present a robust approach to defect identification within the NDI process, leveraging state-of-the-art image segmentation techniques. The proposed methods demonstrate a high defect detection rate while minimizing the need for extensive data pre-processing and computational resources. The rapid detection capabilities inherent to these methods empower inspectors in the NDI process and significantly enhance productivity.

By integrating this approach into the defect detection pipeline, we can effectively reduce inspector fatigue, optimize productivity, and elevate the overall safety standards for manufactured parts. This integration can lead to a more efficient and reliable NDI process, reducing the likelihood of human error and improving the quality of the final product.

\section{Acknowledgement} \label{acknowledgement}
This work was supported in part by the U.S. Department of Energy, Office of Science, under contract \textbf{DE-AC02- 06CH11357}.

\bibliography{references}

\begin{thebibliography}{41}
\providecommand{\natexlab}[1]{#1}

\bibitem[{Baumgartl et~al.(2020)Baumgartl, Tomas, Buettner, and Merkel}]{Baumgartl2020-fe}
Baumgartl, H.; Tomas, J.; Buettner, R.; and Merkel, M. 2020.
\newblock {A deep learning-based model for defect detection in laser-powder bed fusion using in-situ thermographic monitoring}.
\newblock \emph{Progress in Additive Manufacturing}, 5(3): 277--285.

\bibitem[{Brosnan and Sun(2004)}]{Brosnan2004-yd}
Brosnan, T.; and Sun, D.-W. 2004.
\newblock {Improving quality inspection of food products by computer vision----a review}.
\newblock \emph{J. Food Eng.}, 61(1): 3--16.

\bibitem[{Cheng et~al.(2020)Cheng, Collins, Zhu, Liu, Huang, Adam, and Chen}]{Cheng2020-sa}
Cheng, B.; Collins, M.~D.; Zhu, Y.; Liu, T.; Huang, T.~S.; Adam, H.; and Chen, L.-C. 2020.
\newblock {Panoptic-DeepLab}: A simple, strong, and fast baseline for bottom-up panoptic segmentation.
\newblock In \emph{2020 {IEEE/CVF} Conference on Computer Vision and Pattern Recognition ({CVPR})}. IEEE.

\bibitem[{Diamanti, Soutis, and Hodgkinson(2005)}]{Diamanti2005-mr}
Diamanti, K.; Soutis, C.; and Hodgkinson, J.~M. 2005.
\newblock {Non-destructive inspection of sandwich and repaired composite laminated structures}.
\newblock \emph{Compos. Sci. Technol.}, 65(13): 2059--2067.

\bibitem[{Dwivedi, Vishwakarma, and Soni(2018)}]{Dwivedi2018-ts}
Dwivedi, S.~K.; Vishwakarma, M.; and Soni, P.~A. 2018.
\newblock Advances and Researches on Non Destructive Testing: A Review.
\newblock \emph{Materials Today: Proceedings}, 5(2, Part 1): 3690--3698.

\bibitem[{Elharrouss et~al.(2021)Elharrouss, Al-Maadeed, Subramanian, Ottakath, Almaadeed, and Himeur}]{Elharrouss2021-wa}
Elharrouss, O.; Al-Maadeed, S.; Subramanian, N.; Ottakath, N.; Almaadeed, N.; and Himeur, Y. 2021.
\newblock Panoptic Segmentation: A Review.

\bibitem[{Gardner et~al.(2020)Gardner, Fuentes, Dervilis, Mineo, Pierce, Cross, and Worden}]{Gardner2020-bn}
Gardner, P.; Fuentes, R.; Dervilis, N.; Mineo, C.; Pierce, S.~G.; Cross, E.~J.; and Worden, K. 2020.
\newblock {Machine learning at the interface of structural health monitoring and non-destructive evaluation}.
\newblock \emph{Philos. Trans. A Math. Phys. Eng. Sci.}, 378(2182): 20190581.

\bibitem[{Gholizadeh(2016)}]{Gholizadeh2016-ww}
Gholizadeh, S. 2016.
\newblock A review of non-destructive testing methods of composite materials.
\newblock \emph{Procedia Structural Integrity}, 1: 50--57.

\bibitem[{Gupta et~al.(2022)Gupta, Khan, Butola, and Singari}]{Gupta2022-id}
Gupta, M.; Khan, M.~A.; Butola, R.; and Singari, R.~M. 2022.
\newblock {Advances in applications of Non-Destructive Testing (NDT): A review}.
\newblock \emph{Advances in Materials and Processing Technologies}, 8(2): 2286--2307.

\bibitem[{Hafiz and Bhat(2020)}]{Hafiz2020-id}
Hafiz, A.~M.; and Bhat, G.~M. 2020.
\newblock {A survey on instance segmentation: state of the art}.
\newblock \emph{Int. J. Multimed. Inf. Retr.}, 9(3): 171--189.

\bibitem[{Hassani and Dackermann(2023)}]{Hassani2023-hi}
Hassani, S.; and Dackermann, U. 2023.
\newblock {A Systematic Review of Advanced Sensor Technologies for Non-Destructive Testing and Structural Health Monitoring}.
\newblock \emph{Sensors}, 23(4).

\bibitem[{Honarvar and Varvani-Farahani(2020)}]{Honarvar2020-tw}
Honarvar, F.; and Varvani-Farahani, A. 2020.
\newblock A review of ultrasonic testing applications in additive manufacturing: Defect evaluation, material characterization, and process control.
\newblock \emph{Ultrasonics}, 108: 106227.

\bibitem[{Irfan et~al.(2021)Irfan, Almazroi, Rauf, Dama{\v s}evi{\v c}ius, Nasr, and Abdelgawad}]{Irfan2021-ra}
Irfan, R.; Almazroi, A.~A.; Rauf, H.~T.; Dama{\v s}evi{\v c}ius, R.; Nasr, E.~A.; and Abdelgawad, A.~E. 2021.
\newblock Dilated Semantic Segmentation for Breast Ultrasonic Lesion Detection Using Parallel Feature Fusion.
\newblock \emph{Diagnostics (Basel)}, 11(7).

\bibitem[{Jocher, Jing, and Chaurasia(2023)}]{Jocher2023-xo}
Jocher, G.; Jing, Q.; and Chaurasia, A. 2023.
\newblock {Ultralytics {YOLO}}.

\bibitem[{Jodhani et~al.(2023)Jodhani, Handa, Gautam, {Ashwni}, and Rana}]{Jodhani2023-mn}
Jodhani, J.; Handa, A.; Gautam, A.; {Ashwni}; and Rana, R. 2023.
\newblock {Ultrasonic non-destructive evaluation of composites: A review}.
\newblock \emph{Materials Today: Proceedings}, 78: 627--632.

\bibitem[{Khanam and Hussain(2024)}]{Khanam2024-hp}
Khanam, R.; and Hussain, M. 2024.
\newblock {YOLOv11: An overview of the key architectural enhancements}.
\newblock \emph{arXiv [cs.CV]}.

\bibitem[{Khedmatgozar~Dolati et~al.(2021)Khedmatgozar~Dolati, Caluk, Mehrabi, and Khedmatgozar~Dolati}]{Khedmatgozar_Dolati2021-as}
Khedmatgozar~Dolati, S.~S.; Caluk, N.; Mehrabi, A.; and Khedmatgozar~Dolati, S.~S. 2021.
\newblock {Non-Destructive Testing Applications for Steel Bridges}.
\newblock \emph{NATO Adv. Sci. Inst. Ser. E Appl. Sci.}, 11(20): 9757.

\bibitem[{Kirillov et~al.(2019)Kirillov, He, Girshick, Rother, and Dollar}]{Kirillov2019-uc}
Kirillov, A.; He, K.; Girshick, R.; Rother, C.; and Dollar, P. 2019.
\newblock Panoptic Segmentation.
\newblock In \emph{2019 {IEEE/CVF} Conference on Computer Vision and Pattern Recognition ({CVPR})}. IEEE.

\bibitem[{Lange(1994)}]{Lange1994-nf}
Lange, Y.~V. 1994.
\newblock {THE MECHANICAL IMPEDANCE ANALYSIS METHOD OF NONDESTRUCTIVE TESTING (A REVIEW)}.
\newblock \emph{NDT E Int.}, 11(4): 177--193.

\bibitem[{Li, Wang, and Liu(2024)}]{li2024yolo}
Li, H.; Wang, C.; and Liu, Y. 2024.
\newblock YOLO-FDD: efficient defect detection network of aircraft skin fastener.
\newblock \emph{Signal, Image and Video Processing}, 18(4): 3197--3211.

\bibitem[{Lin et~al.(2014)Lin, Maire, Belongie, Bourdev, Girshick, Hays, Perona, Ramanan, Lawrence~Zitnick, and Doll{\'a}r}]{Lin2014-jw}
Lin, T.-Y.; Maire, M.; Belongie, S.; Bourdev, L.; Girshick, R.; Hays, J.; Perona, P.; Ramanan, D.; Lawrence~Zitnick, C.; and Doll{\'a}r, P. 2014.
\newblock Microsoft {COCO}: Common Objects in Context.

\bibitem[{Lin et~al.(2023)Lin, Ma, Wang, and Sun}]{Lin2023-fd}
Lin, Y.; Ma, J.; Wang, Q.; and Sun, D.-W. 2023.
\newblock {Applications of machine learning techniques for enhancing nondestructive food quality and safety detection}.
\newblock \emph{Crit. Rev. Food Sci. Nutr.}, 63(12): 1649--1669.

\bibitem[{Ma, Tsuchikawa, and Inagaki(2020)}]{Ma2020-cu}
Ma, T.; Tsuchikawa, S.; and Inagaki, T. 2020.
\newblock {Rapid and non-destructive seed viability prediction using near-infrared hyperspectral imaging coupled with a deep learning approach}.
\newblock \emph{Comput. Electron. Agric.}, 177: 105683.

\bibitem[{Memmolo et~al.(2015)Memmolo, Arena, Fatigati, Grilli, Paturzo, Pezzati, and Ferraro}]{Memmolo2015-er}
Memmolo, P.; Arena, G.; Fatigati, G.; Grilli, M.; Paturzo, M.; Pezzati, L.; and Ferraro, P. 2015.
\newblock {Automatic Frames Extraction and Visualization From Noisy Fringe Sequences for Data Recovering in a Portable Digital Speckle Pattern Interferometer for NDI}.
\newblock \emph{J. Display Technol.}, 11(5): 417--422.

\bibitem[{Mishra, Bhatia, and Maity(2020)}]{Mishra2020-qv}
Mishra, M.; Bhatia, A.~S.; and Maity, D. 2020.
\newblock {Predicting the compressive strength of unreinforced brick masonry using machine learning techniques validated on a case study of a museum through nondestructive testing}.
\newblock \emph{Journal of Civil Structural Health Monitoring}, 10(3): 389--403.

\bibitem[{Ould~Naffa et~al.(2002)Ould~Naffa, Goueygou, Piwakowski, and Buyle-Bodin}]{Ould_Naffa2002-ab}
Ould~Naffa, S.; Goueygou, M.; Piwakowski, B.; and Buyle-Bodin, F. 2002.
\newblock {Detection of chemical damage in concrete using ultrasound}.
\newblock \emph{Ultrasonics}, 40(1-8): 247--251.

\bibitem[{Prakash et~al.(2023)Prakash, Nieberl, Mayer, and Schuster}]{prakash2023learning}
Prakash, N.; Nieberl, D.; Mayer, M.; and Schuster, A. 2023.
\newblock Learning defects from aircraft NDT data.
\newblock \emph{NDT \& E International}, 138: 102885.

\bibitem[{Rahman and Wang(2016)}]{Rahman2016-uu}
Rahman, M.~A.; and Wang, Y. 2016.
\newblock {Optimizing intersection-over-union in deep neural networks for image segmentation}.
\newblock In \emph{{Advances in Visual Computing}}, Lecture notes in computer science, 234--244. Cham: Springer International Publishing.
\newblock ISBN 9783319508344,9783319508351.

\bibitem[{Redmon et~al.(2016)Redmon, Divvala, Girshick, and Farhadi}]{Redmon2016-ap}
Redmon, J.; Divvala, S.; Girshick, R.; and Farhadi, A. 2016.
\newblock {You only look once: Unified, real-time object detection}.
\newblock In \emph{{2016 IEEE Conference on Computer Vision and Pattern Recognition (CVPR)}}, 779--788. IEEE.
\newblock ISBN 9781467388511,9781467388528.

\bibitem[{Su et~al.(2011)Su, Wang, Jiao, and Guo}]{Su2011-re}
Su, Y.; Wang, Y.; Jiao, J.; and Guo, Y. 2011.
\newblock Automatic detection and classification of breast tumors in ultrasonic images using texture and morphological features.
\newblock \emph{Open Med. Inform. J.}, 5(Suppl 1): 26--37.

\bibitem[{Sun, Ramuhalli, and Jacob(2023)}]{Sun2023-yp}
Sun, H.; Ramuhalli, P.; and Jacob, R.~E. 2023.
\newblock {Machine learning for ultrasonic nondestructive examination of welding defects: A systematic review}.
\newblock \emph{Ultrasonics}, 127: 106854.

\bibitem[{Taheri, Gonzalez~Bocanegra, and Taheri(2022)}]{Taheri2022-ix}
Taheri, H.; Gonzalez~Bocanegra, M.; and Taheri, M. 2022.
\newblock {Artificial Intelligence, Machine Learning and Smart Technologies for Nondestructive Evaluation}.
\newblock \emph{Sensors}, 22(11).

\bibitem[{Thomas et~al.(2024)Thomas, Kuang, Wang, Barnes, and Jenkins}]{thomas2024advanced}
Thomas, J.; Kuang, B.; Wang, Y.; Barnes, S.; and Jenkins, K. 2024.
\newblock Advanced semantic segmentation of aircraft main components based on transfer learning and data-driven approach.
\newblock \emph{The Visual Computer}, 1--20.

\bibitem[{{Ultralytics}(2023)}]{Ultralytics2023-lc}
{Ultralytics}. 2023.
\newblock {Object Detection Datasets Overview}.
\newblock \url{https://docs.ultralytics.com/datasets/detect}.
\newblock Accessed: 2024-11-21.

\bibitem[{{Ultralytics}(2024)}]{Ultralytics2024-em}
{Ultralytics}. 2024.
\newblock {YOLO11 {NEW}}.
\newblock \url{https://docs.ultralytics.com/models/yolo11}.
\newblock Accessed: 2024-11-21.

\bibitem[{Wang et~al.(2020)Wang, Zhong, Lee, Fancey, and Mi}]{Wang2020-up}
Wang, B.; Zhong, S.; Lee, T.-L.; Fancey, K.~S.; and Mi, J. 2020.
\newblock {Non-destructive testing and evaluation of composite materials/structures: A state-of-the-art review}.
\newblock \emph{Advances in Mechanical Engineering}, 12(4): 1687814020913761.

\bibitem[{Wu et~al.(2019{\natexlab{a}})Wu, Kirillov, Massa, Lo, and Girshick}]{wu2019detectron2}
Wu, Y.; Kirillov, A.; Massa, F.; Lo, W.-Y.; and Girshick, R. 2019{\natexlab{a}}.
\newblock Detectron2.
\newblock \url{https://github.com/facebookresearch/detectron2}.

\bibitem[{Wu et~al.(2019{\natexlab{b}})Wu, Kirillov, Massa, Lo, and Girshick}]{Wu2019-iw}
Wu, Y.; Kirillov, A.; Massa, F.; Lo, W.-Y.; and Girshick, R. 2019{\natexlab{b}}.
\newblock Detectron2.

\bibitem[{Xiang, Chang, and Ye(2024)}]{xiang2024aircraft}
Xiang, B.; Chang, H.; and Ye, Y. 2024.
\newblock Aircraft Part Recognition from Drones’ Perspective Based on Semantic Segmentation.
\newblock In \emph{2024 IEEE 7th International Conference on Electronic Information and Communication Technology (ICEICT)}, 1288--1292. IEEE.

\bibitem[{Xiong et~al.(2019)Xiong, Liao, Zhao, Hu, Bai, Yumer, and Urtasun}]{Xiong2019-qk}
Xiong, Y.; Liao, R.; Zhao, H.; Hu, R.; Bai, M.; Yumer, E.; and Urtasun, R. 2019.
\newblock {UPSNet}: A Unified Panoptic Segmentation Network.
\newblock In \emph{2019 {IEEE/CVF} Conference on Computer Vision and Pattern Recognition ({CVPR})}. IEEE.

\bibitem[{Yang et~al.(2022)Yang, Gao, Chen, Yuan, Chen, and Kong}]{Yang2022-nl}
Yang, Z.; Gao, W.; Chen, L.; Yuan, C.; Chen, Q.; and Kong, Q. 2022.
\newblock {A novel electromechanical impedance-based method for non-destructive evaluation of concrete fiber content}.
\newblock \emph{Construction and Building Materials}, 351: 128972.

\end{thebibliography}

\end{document}